\RequirePackage{amsmath}
\documentclass[runningheads]{llncs}

\usepackage[T1]{fontenc}

\usepackage{graphicx}

\usepackage{bbding}

\usepackage{hyperref}
\usepackage{color}

\usepackage{amssymb}
\usepackage{amsmath}
\usepackage{bm}

\usepackage{booktabs}
\usepackage{multirow}
\usepackage{tabularx}
\usepackage{makecell}
\newcolumntype{Y}{>{\centering\arraybackslash}X}
\usepackage{threeparttable}

\usepackage{pifont}
\newcommand{\vmark}{\ding{51}}
\newcommand{\xmark}{\ding{55}}

\usepackage[inline]{enumitem}

\begin{document}

\title{Relaxed syntax modeling in Transformers for~future-proof license plate recognition}
\titlerunning{Relaxed syntax modeling in Transformers for future-proof LPR}

\author{Florent Meyer\inst{1,2}~\Envelope~~\orcidID{0009-0001-7527-5215} \and
Laurent Guichard\inst{1}\orcidID{0009-0008-7853-8704} \and
Denis Coquenet\inst{2}\orcidID{0000-0001-5203-9423} \and
Guillaume Gravier\inst{2}\orcidID{0000-0002-2266-5682} \and
Yann Soullard\inst{3}\orcidID{0009-0001-8048-2489} \and
Bertrand Coüasnon\inst{2}\orcidID{0000-0002-7077-0751}}
\authorrunning{F. Meyer et al.}

\institute{ANTAI, Rennes, France \and
Univ Rennes, CNRS, IRISA - UMR 6074, Rennes, France \and
Univ Rennes, Université Rennes 2, CNRS, IRISA - UMR 6074, Rennes, France
\\
\email{florent.meyer@irisa.fr}}

\maketitle

\begin{abstract}
Effective license plate recognition systems are required to be resilient to constant change, as new license plates are released into traffic daily. While Transformer-based networks excel in their recognition at first sight, we observe significant performance drop over time which proves them unsuitable for tense production environments. Indeed, such systems obtain state-of-the-art results on plates whose syntax is seen during training. Yet, we show they perform similarly to random guessing on future plates where legible characters are wrongly recognized due to a shift in their syntax. After highlighting the flows of positional and contextual information in Transformer encoder-decoders, we identify several causes for their over-reliance on past syntax. Following, we devise architectural cut-offs and replacements which we integrate into SaLT, an attempt at a Syntax-Less Transformer for syntax-agnostic modeling of license plate representations. Experiments on both real and synthetic datasets show that our approach reaches top accuracy on past syntax and most importantly nearly maintains performance on future license plates. We further demonstrate the robustness of our architecture enhancements by way of various ablations.

\end{abstract}

\setcounter{footnote}{0}

\section{\label{intro}Introduction}
Vehicle License Plate Recognition (LPR) has been in use for several years in many countries with applications ranging from traffic surveillance to speed limit monitoring. In the context of traffic law enforcement notably, the tremendous volume of contraventions recorded daily needs efficient automatic recognition so as to minimize the amount of human workload. Consequently, LPR has benefitted from gradual advances in deep learning research for vision and state-of-the-art results are now obtained with Transformers~\cite{kumarDPAMNewDeep2022,moussaForensicLicensePlate2022,xueImagetoCharactertoWordTransformersAccurate2023}.

\begin{table}
\caption{\textbf{Exact match on real LP of TrOCR with diverse fine-tuning, pretraining and tokenization options.} TrOCR consistently fails at decoding images of future LP starting with a \texttt{G} (target syntax), whether the whole model is fine-tuned or not. Fine-tuning only enables reaching decent whole-LP accuracy on plates starting with letters \texttt{A} to \texttt{F} (source syntax) when a character-level tokenizer replaces byte-pair-encoding (BPE)~\cite{sennrichNeuralMachineTranslation2016}. \textit{stage1} denotes the first-stage pretrained checkpoint from~\cite{liTrOCRTransformerbasedOptical2023}.}
\label{table_frlpr_fail_trocr}
\centering
\setlength{\tabcolsep}{4pt}
\begin{tabular}{cccccc}
\toprule
\multirow{2}[4]{*}{\textbf{Tokenizer}} & \multicolumn{2}{c}{\textbf{Pretraining}}                   & \multirow{2}[4]{*}{\textbf{Fine-tuning}} & {\textbf{\makecell{Target \\ syntax}}} & {\textbf{\makecell{Source \\ syntax}}} \\ \cmidrule(lr){2-3} 
                                    & Part                             & Checkpoint              &                                       & {\verb/^G.*$/}                                                    & \verb/^[A-F].*$/                                                    \\ \midrule
\multirow{2}{*}{BPE}                & \multirow{2}{*}{Encoder-decoder} & stage1                  & \multirow{2}{*}{\xmark}                                   & 1.7                                                 & 1.3                                                 \\
                                    &                                  & printed                 &                                                           & 3.8                                                 & 3.0                                                 \\ \hline \\[-2ex]
BPE                                 & Encoder-decoder                  & \multirow{2}{*}{stage1} & \multirow{2}{*}{\vmark}                                   & $0.9\pm1.7$                                             & $99.4\pm0.0$                                             \\
Character                           & Encoder                          &                         &                                                           & $27.8\pm32.8$                                       & $99.5\pm0.0$                                        \\ \bottomrule
\end{tabular}
\end{table}

\begin{figure}
\centering
\includegraphics[width=0.95\textwidth]{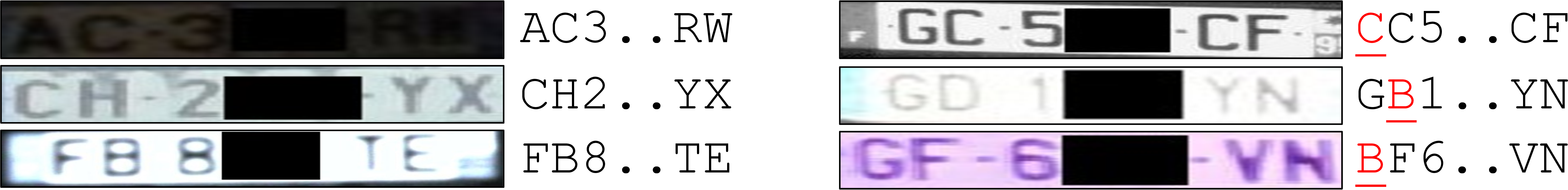}
\caption{\textbf{Cropped photographs of real test LP with predictions by TrOCR.} \textbf{Left:} Fine-tuned TrOCR successfully decodes images of LP starting with letters from \texttt{A} to \texttt{F}, (training-time syntax), despite degraded quality. \textbf{Right:} Yet, it fails on legible LP with a \texttt{G} in the leftmost position (future syntax). Errors (underlined) occur mainly in the first and second positions. LP are anonymised for RGPD compliance.}
\label{fig_frlpr_fail_trocr}
\end{figure}

Most LPR systems are evaluated on test datasets whose distribution is practically identical to that of the training samples, thereby reaching satisfying performance and seeming ready to be run in production. Despite remarkable improvements in recognition accuracy, there remain challenges to be addressed. Indeed, unforeseen difficulties may arise from a misrepresentation of the syntax followed by license plates (LP) on the roads, for it is different from that learned by a deep network~\cite{larocaEfficientLayoutIndependentAutomatic2021,resendegoncalvesRealTimeAutomaticLicense2018}. For instance, some countries use serials with characters increased by an increment from the right with each LP put in service. This specific scheme regularly leads to a tipping point, namely a \textit{shift}, when the character updated the least often (e.g. in the leftmost position) is eventually increased as well. Put another way, a known character previously missing in a given position now appears in it. From then on, the content of the training dataset learned by the model will be less and less representative of the real-life LP in circulation as an increasing number of new-syntax LP make their way into traffic. Throughout this paper, we report the non-negligible impact this change in syntax has on cropped LP transcription accuracy of Transformer-based architectures. As showcased in Table~\ref{table_frlpr_fail_trocr}, TrOCR~\cite{liTrOCRTransformerbasedOptical2023} largely underperforms on target syntax beginning with letter \texttt{G} if LP starting with letters from \texttt{A} to \texttt{F} are seen during training. Note that off-the-shelf checkpoints are not specifically targeted at LPR. Fig.~\ref{fig_frlpr_fail_trocr} depicts typical cases of success and failure in TrOCR's predictions. While we focus on LPR in this paper, similar syntax shift can be observed in regular expression-based use cases like ID, date, invoice or serial number recognition.

Starting from a Transformer encoder-decoder widely used for text recognition, we try to understand the root of its memorization of source syntax, i.e. that seen at training time. By studying the flows of contextual and positional information within the network, we identify several causes of inconvenient memorization both in the encoder and decoder parts. Based on these observations, we propose the Syntax-Less Transformer (SaLT), a light Transformer-based encoder-decoder equipped with elements devised to preserve performance over time without retraining. Indeed, retraining is not satisfactory because the in-production behavior would be unpredictable and often erroneous on plates with new syntax until enough such images are collected, annotated and used for retraining. SaLT is derived from a \textit{debiasing} framework with a convolutional encoder and enhancements regarding the input and cross-attention of the Transformer-decoder. Extensive evaluation reveals SaLT's robustness to a constantly evolving syntax, bringing a large gain on target syntax while strongly reducing performance variability. Meanwhile, performance on source syntax serials is fully retained. Additional experiments are conducted on LPR-MNIST, a new synthetic dataset which replicates LP syntax evolution through time, publicly released for future experiments on syntax evolution\footnote{Available at \url{https://www-shadoc.irisa.fr/lpr-mnist-dataset/}.}. Finally, ablation studies reveal the combined effect of our modifications on robustness to syntax shift.

\section{Related work}

\subsection{License plate recognition}
With applications from traffic monitoring to parking fee payment, LPR systems have benefited from advances in deep learning research over the years. In this work, we consider images cropped by a localization module and resized around the LP beforehand, and thus focus on text recognition. Early multi-step architectures were generally composed of separate processing methods, typically segmenting plate characters before recognition of each individually~\cite{anagnostopoulosLicensePlateRecognition2008}. End-to-end deep networks appeared later on, notably with object-detection methods derived from YOLO~\cite{montazzolliRealTimeBrazilianLicense2017} detecting each character class separately. In our case however, neither character-level bounding box nor segmentation annotations are available due to their high labelling cost, we thus dismiss related methods. Indeed, we only have plate-level transcriptions. Moreover, the text on some LP is shared between two lines, e.g. on motorbikes. Thus, one-dimensional sequence alignement methods like connectionist temporal classification~\cite{gravesConnectionistTemporalClassification2006} are unsuited. Long short-term memory (LSTM)-based sequence modeling~\cite{zhangRobustAttentionalFramework2021} was then applied to LPR. More recently, Transformers have gained interest in research although not widely applied to LPR yet~\cite{fuDeepLearningLicense2024,khanLicensePlateRecognition2023}, outperforming recurrent networks like LSTM. For instance in~\cite{moussaForensicLicensePlate2022,xueImagetoCharactertoWordTransformersAccurate2023}, existing architectures are modified by inputting the image compression level to improve recognition under strong compression or by decomposing text recognition into two inter-connected tasks, namely image-to-character and character-to-word mapping. Among publicly available off-the-shelf models, the pure Transformer TrOCR~\cite{liTrOCRTransformerbasedOptical2023} shows promising results for LPR \cite{raiTrALPRAutomaticLicense2023,wuEmissionAnalysisBased2024}.

\subsection{Remediation to syntax shift}
Diverse approaches attempt to increase robustness to a shifted target syntax, notably data augmentation, training strategy and additional model components. 

\subsubsection{Data augmentation} 
One family of methods aims at obtaining a balanced training dataset, altering neither the model nor the learning process. The authors of~\cite{resendegoncalvesRealTimeAutomaticLicense2018} control the frequency of each character on LP by permuting them, swapping overrepresented letters with underrepresented ones to increase their ratio. Similarly in \cite{yimSynthTIGERSyntheticText2021}, infrequent characters are augmented by image synthesis with diverse backgrounds, font styles, and text shapes. Alternatively, in order to mitigate the imbalance between plates of diverse countries, Laroca et al.~\cite{larocaEfficientLayoutIndependentAutomatic2021} opt for gathering and manually labeling more LP photographs from the Internet, which is a costly solution. To adapt to new languages and overcome an everchanging data distribution in multilingual scene text recognition (STR), an ensemble method of a language identifier followed by one module per language was devised in~\cite{zhengMRNMultiplexedRouting2023}. This requires retraining for each newly supported language. 
Conversely, we choose not to augment our dataset with artificial LP of all imaginable future character combinations. Rather than depend upon the data itself, we explore a more architecture-centered approach to debiasing without retraining. 

\subsubsection{Training strategy \& additional components} 
Another family of approaches deal with syntax shift in STR through sophisticated training procedures or additional network components. On the one hand, some methods suppose the availability of target syntax samples at training time. The authors of~\cite{parkImprovingSceneText2023} explore an ensemble method to train on a long-tailed character distribution while preserving performance on a balanced dataset. Two separate experts learn on the unbalanced source dataset and a smaller target dataset with a balanced number of characters, respectively. So as to reduce bias towards high-frequency characters, Tran Tien et al.~\cite{trantienUnsupervisedDomainAdaptation2023} use unlabeled target data to minimize the latent entropy of the predicted probability distribution over characters. By investigating the effect of unidirectional language priors, \cite{huVisionLanguageAdaptiveMutual2023} propose an adaptive fusion module and add Kullback-Leibler divergence losses between left-to-right and right-to-left decoding passes. These methods are inapplicable to our use case on account of the temporal evolution of LP syntax, as no target syntax samples are available at training time. On the other hand, some contrastive or mutual learning approaches work without any real target syntax samples. In order to alleviate over-fitting due to lexical dependencies in multiple STR tasks, it was proposed to randomly permute scene text image patches in a contrastive learning fashion, with each patch representing several characters~\cite{zhangRelationalContrastiveLearning2023,zhangContextBasedContrastiveLearning2022}. Another way of reducing bias towards source vocabulary is a mutual learning strategy between two complementary models with additional Kullback-Leibler divergence loss between distributions predicted by final layers~\cite{wanVocabularyRelianceScene2020a}. As opposed to these methods which increase model size with additional modules or complicate training, Liu et al.~\cite{liuImportanceWordOrder2021} reduce word order information for cross-lingual sequence labeling by replacing positional encoding with convolutions after Transformer attention. Liu et al.~\cite{liuImprovingZeroShotTranslation2021} disentangle positional information to improve zero-shot translation by removing residual connections in a Transformer-encoder. In this way, we propose architectural enhancements inspired by information flow analysis that leave the training strategy unchanged and do not involve any extra model parameters.

\begin{figure}
\centering
\includegraphics[width=0.87\textwidth]{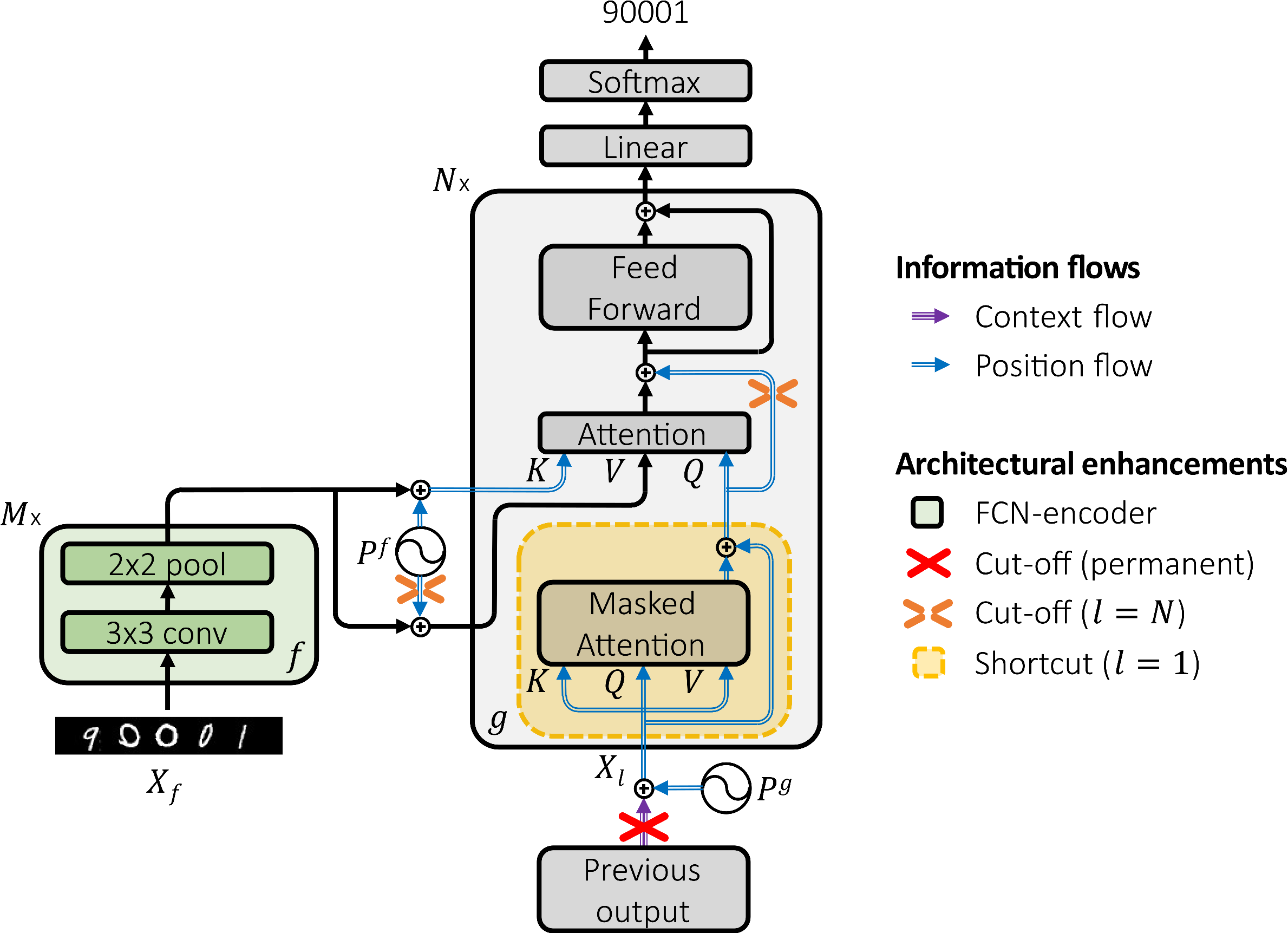}
\caption{\textbf{Overall debiasing framework applied to a Transformer encoder-decoder network}. The proposed architectural enhancements discard positional and contextual bias through an accurate control of information flows.} \label{fig_salt}
\end{figure}

\section{Debiasing framework}
We start by formulating the syntax shift problem and the two resulting roots of bias in Transformers. Second, we formalize the advantage of preferring a fully convolutional encoder over a Transformer one for successful target syntax recognition. Third, we suggest syntax-debiasing enhancements to a vanilla Transformer-decoder. We finally assemble these two parts into SaLT, a debiased encoder-decoder architecture. The overall framework is illustrated in Fig.~\ref{fig_salt}.

\subsection{Context \& problem statement}
Transformer encoder-decoders have demonstrated their ability to exceed the performance of previous architectures in LPR~\cite{moussaForensicLicensePlate2022,wuEmissionAnalysisBased2024,xueImagetoCharactertoWordTransformersAccurate2023,zhangAutomaticLicensePlate2023}. Notably, \cite{raiTrALPRAutomaticLicense2023} achieves promising results using TrOCR~\cite{liTrOCRTransformerbasedOptical2023}. We choose it as a baseline since it is a vanilla Transformer with minimal adaptations for STR. It can therefore be seen as a representative of Transformers and their bias. Albeit displaying excellent fine-tuned performance on a source syntax dataset, TrOCR fails to reliably recognize target syntax as shown in Table~\ref{table_frlpr_fail_trocr}. The network obviously cannot read known characters in new positions on an LP. On the contrary, we argue it learns a biased strategy preventing it from achieving decent performance on future data. 

Let us now consider the differences between source and target data which may lead to poor reading of target syntax. Let $\Omega$ be the set of all character classes. For each position, we define $C^S_i$ and $C^T_i$ as the sets of characters that can be found in position $i$ for source and target syntax, respectively. One can define the source syntax $\mathcal{S}$ and the target syntax $\mathcal{T}$ as:
\begin{equation}
\begin{aligned}
\mathcal{S} = \{c_1,  ..., c_L \; | \; c_i \in \mathcal{C}^\text{S}_i, \, \mathcal{C}^\text{S}_i \subseteq \Omega\}  \\
\mathcal{T} = \{c_1,  ..., c_L \; | \; c_i \in \mathcal{C}^\text{T}_i, \, \mathcal{C}^\text{T}_i \subseteq \Omega\} 
\end{aligned}
\end{equation}
where $L$ is the maximum number of characters on a LP. If training and evaluation are performed on plates from the same time period, $T = S$. In our case however, $\exists \, i, \, \mathcal{C}^\text{T}_i = \mathcal{C}^\text{S}_i \cup \mathcal{N}_i$ where $\mathcal{N}_i$ is a non-empty set of new characters appeared after training, hence $T \neq S$. To correspond to LP evolution, we mainly focus on the case where the possible positions of characters in the sequence are the same for all but the leftmost one, i.e. $\mathcal{C}^\text{T}_1 = \mathcal{C}^\text{S}_1 \cup \mathcal{N}_1$. Visually, some letters or digits never appear in the leftmost position on the image, closest to the country indicator. Textually, this translates into a lack of examples of said characters as first in the sequence. Examples of this shift are illustrated in Fig.~\ref{fig_frlpr_fail_trocr}. 

We assume the drop in performance of Transformer encoder-decoder architectures results from them memorizing aforementioned particularities in training data. Henceforth, we hypothesize such networks have several architectural mechanisms causing them to learn undesirable data bias which we respectively name \textit{contextual} and \textit{positional} bias. Yet, we suspect our real LP dataset to contain relevant information for creating robust latent representations of individual characters which could be leveraged for recognition in positions in which they never occurred at training time. Therefore, we now detail how the encoder and decoder can each learn both types of bias in their own way, and how to alleviate it. 

\subsection{\label{encoder}Vision encoder}
Our first proposal is to replace the vision Transformer-encoder with a fully convolutional network (FCN) denoted by $f$.
Inspired by recent text recognition encoder-decoder architectures~\cite{coquenetDANSegmentationfreeDocument2023,singhFullPageHandwriting2021}, we use a convolutional, attention-less encoder in order to control its receptive field and discard positional information.

\subsubsection{Visual positional bias}
Positional bias is intrinsically linked to the Transformer architecture, more particularly to the multi-head self-attention operation included in each encoder layer. Indeed, due to the permutation-invariance property of attention, it is necessary to inject explicit positional information to its input using positional encoding~\cite{vaswaniAttentionAllYou2017}. In this way, we argue Transformer-encoders bind absolute position information to character class. Let us recall that attention computes a weighted sum of values $\bm{V}$ according to attention weights dot-product from queries $\bm{Q}$ and keys $\bm{K}$. Specifically, if absolute position information is tied to encoded features sent as $\bm{V}$ for cross-attention, then it will impact the decoder's prediction. Thus, if some character is absent from a given zone in all training images then it can hardly be predicted in this zone. For instance, the model can learn that no \texttt{G} should be predicted near the left edge of a LP.

An FCN instead processes images and outputs features without position information.

\subsubsection{Visual contextual bias}
On the one hand, except for attempts at benefiting from so-called \textit{local inductive bias}~\cite{dascoliConViTImprovingVision2022,liuSwinTransformerHierarchical2021}, each vanilla Transformer layer with attention can~\textit{see the whole picture}. That is, its field of vision is not limited to a local region of the image, which allows it to learn correlations between elements located anywhere. When combined with above-mentioned mandatory position information, we suspect this leads to learning strong visual biases, among which inter-character relations, proximity with a country indicator, plate border, space or dash, etc. On the other hand, it is widely held that it is difficult for convolutional networks to capture long-range relationships, especially when using small filter kernels as in popular networks like ResNet~\cite{heDeepResidualLearning2016}. Indeed, the receptive field of an FCN slowly increases along layers without necessarily covering the whole image at the end of the network.

As a countermeasure to visual context modeling, we thus plainly employ convolution blocks aiming for a final receptive field $r$ of approximately the size of a character at the output of the encoder $f$. The main goal here is to disentangle the latent representations of successive characters so that features of a given character do not depend on the adjacent ones. It should not be large enough to cover large regions of the image and thereby absorb contextual bias, neither should it be too small, as feature modeling could otherwise be negatively impacted.

\subsection{\label{decoder}Text decoder}
Sources of both contextual and positional bias are expected to be found in a Transformer-decoder too, denoted by $g$. However, we cannot totally get rid of this component as it is the best suited for multi-line text recognition. Hence, we bring changes to the cross-attention block and to the input of the decoder in order to eliminate syntax modeling and restrain the flow of positional information. 

\subsubsection{Textual contextual bias} 
In autoregressive Transformer-decoders~\cite{vaswaniAttentionAllYou2017}, left-to-right (LTR) language modeling emerges from the teacher forcing method. It consists in training the model to output the next text occurrence at each time step for a given sample using its left-side context while right-side tokens are masked. Let us recall that such samples begin with a special start-of-sentence token \texttt{<s>} and that absolute positional encoding is added to all token embeddings for the model to know the ordering of the input tokens. This suggests that a character, say \texttt{G}, appearing in a new position, say the leftmost one, would disrupt decoding for two reasons. First, $\mathbb{P} (\texttt{G}|\texttt{<s>})$ is poorly estimated because it goes against the language rules memorized during training since \texttt{<s>} is an unseen left context. Second, the same goes for $\mathbb{P} (x|\texttt{<s>G})$, where $x \in \mathcal{C}^\text{T}_2 = \mathcal{C}^\text{S}_2$, assuming \texttt{<s>G} was correctly output. These two failure cases are illustrated in Fig.~\ref{fig_frlpr_fail_trocr}. 

To prevent language modeling and move closer to syntactical independence, we entirely remove text token embeddings from the decoder input. More precisely, neither the embeddings of \texttt{<s>} nor those of previously predicted tokens are reinjected as decoder inputs as is usually done. Instead, we use a sequence of \textit{position queries} $\text{\textbf{P}}^\text{g} = (\text{\textbf{p}}_1,...,\text{\textbf{p}}_L)$ containing positional encoding only, where $\text{\textbf{p}}_i \in R^{d}$ refers to a positional encoding vector from the original Transformer~\cite{vaswaniAttentionAllYou2017}. Decoding can thus be parallelized for greater efficiency~\cite{carionEndtoEndObjectDetection2020}, as the decoding process is no longer autoregressive. Also, since position queries are constant vectors, meaning that no relevant information can be extracted from them solely, we short-cut (i.e. remove) self-attention in the first decoder layer. In downstream layers, we keep the usual LTR mask in self-attention as an incentive to sequentially move its gaze on the image. The content of $\bm{Q}$ entering cross-attention can be summarized as follows:
\begin{equation}
\bm{Q}_l=
    \begin{cases}
    \text{\textbf{P}}^\text{g}, & \text{if}\ l=1 \\
    \text{MultiHead}(\bm{X}_l) + \bm{X}_l, & \text{otherwise}
    \end{cases}
\end{equation}
with $\bm{X}_l$ the input to the current decoder layer $l$, $\text{MultiHead}$ the masked multi-head self-attention defined in~\cite{vaswaniAttentionAllYou2017}.

\subsubsection{Positional bias}
So far, we focused on removing inter-character-dependency modeling capabilities. We now interpret the way in which position information \textit{flows} through a standard Transformer-decoder, i.e. how tensors progress along a forward pass in the network. As pictured in Fig.~\ref{fig_salt}, position information enters any decoder layer from two spots: via image features (keys $\bm{K}$ and values $\bm{V}$ of the cross-attention) and position queries $\text{\textbf{P}}^\text{g}$ (decoder input). For the former, the intuition behind stopping position flow is to make cross-attention output a weighted sum of encoded features $\bm{V}$ which do not contain any information about the location inside the text sequence. For the latter, we focus on the residual connection to get rid of the position information before prediction takes place. We take action upon said spots in the last decoder layer only, as it is closest to the final decision making.

\paragraph{Textual bias}
One entry spot is the layer's upstream input. Indeed, we keep previously defined position queries $\text{\textbf{P}}^\text{g}$ because we want cross-attention to sequentially focus on the successive characters in the image with respect to the current position of interest. $\text{\textbf{P}}^\text{g}$ then flows freely from previous layers by taking the residual path over successive cross-attention blocks.

However, we cut off said residual connection in the last layer. We implement this modification in the last layer only to prevent any vanishing gradient problem. Hence, the flow of sequence position information coming from the decoder input is stopped before the prediction layer.

\paragraph{Cross-modal bias}
The other entry of position information into a decoder layer is through the keys $\bm{K}$ and values $\bm{V}$ used in cross-attention. As detailed in Section~\ref{encoder}, the FCN-encoder is guaranteed to provide features $f( \bm{X}_\text{f} )$ which do not contain any position information. Positional encoding must nonetheless be added to image features $f( \bm{X}_\text{f} )$ in order to compute $\bm{K}$. Indeed, cross-modal positioning between image and text is possible only if both modalities, namely $\bm{K}$ and $\bm{Q}$, contain positional information. Let us recall that $\bm{Q}$ already contains $\text{\textbf{P}}^\text{g}$ thanks to the residual path of previous layers.

Yet, as opposed to \cite{coquenetDANSegmentationfreeDocument2023,singhFullPageHandwriting2021}, we choose to leave $\bm{V}$ position-less in the last layer:
\begin{equation}
\bm{K} = f( \bm{X}_\text{f} ) + \text{\textbf{P}}^\text{f} \\ 
\end{equation}
\begin{equation}
\bm{V}_l =
    \begin{cases}
    f( \bm{X}_\text{f} ), & \text{if}\ l = N \\
    f( \bm{X}_\text{f} ) + \text{\textbf{P}}^\text{f}, & \text{otherwise}
    \end{cases}
\end{equation}
where $N$ is the number of decoder layers and $\text{\textbf{P}}^\text{f}$ represents image positional encoding. As such, the cross-attention in the last decoder layer outputs a weighted sum of position-less features $\bm{V}$.

\subsection{\label{modelname}SaLT}
We now propose the Syntax-Less Transformer (SaLT), a custom encoder-decoder architecture depicted in Fig.~\ref{fig_salt} which integrates the key components we have discussed above. Intuitively, its FCN-encoder first provides disentangled character representations, independent of both their visual context and absolute position in the image. Then, regardless of other predicted characters, SaLT's decoder queries each pick among encoded features by searching for those seemingly containing a character and matching the position currently considered. Finally, the decoder prediction is computed from encoded features which do not contain any information about the location inside the image nor the text sequence.

\subsubsection{Fully convolutional encoder}
The encoder takes as input a document image $\bm{X}_\text{f} \in R^{H \times W\times C}$, with $H$, $W$ and $C$ being respectively the height, width, and number of channels. Images of any size can be processed thanks to the FCN-Transformer mix. SaLT's encoder consists in a stack of $M = 4$ convolution blocks. Each block comprises 4 components, namely 
\begin{enumerate*}[label=(\roman*), itemjoin={{, }}, itemjoin*={{, and }}]
\item a $3 \times 3$ convolution with a stride of $2$ which pads in such a way that the output has the same shape as the input
\item instance normalization~\cite{ulyanovImprovedTextureNetworks2017}
\item $2 \times 2$ max pooling 
\item dropout \cite{srivastavaDropoutSimpleWay2014} with a probability of 0.1. 
\end{enumerate*}
This results in a receptive field $r = 46$. In addition, time and memory complexity is greatly reduced compared to a multi-layer Transformer-encoder. The first convolution receives $C_{in}$ channels depending on the color model of the image (grayscale or RGB) and outputs 32 channels. Then, the number of channels doubles at each convolution to reach a final $C_{out} = d = 256$ as expected by the decoder. 

\subsubsection{Debiased Transformer-decoder}
The decoding process of SaLT is handled by a debiased, $N$-layer Transformer decoder with $N=2$. Its internal dimension is $d=256$ and it uses 8 attention heads. Its position-wise 2-layer feed-forward network uses the same internal dimension $d$. SaLT and \textit{debiased} model versions employ two of the three syntax-relaxing decoder modifications, as motivated by the ablation study in Section~\ref{ablation}, namely position queries and residual cut-off. Decoding is done in a parallel fashion.

\section{\label{datasets}Datasets}
In this section, we describe the two datasets at hand, namely a new synthetic LPR-MNIST and the real-life LP dataset. For both datasets, the test split containing both source and target syntax, while samples of target syntax are absent from the training and validation splits. 

\begin{figure}
\centering
\includegraphics[width=\textwidth]{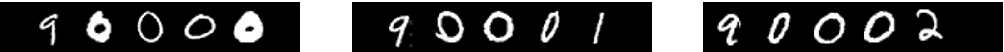}
\caption{\textbf{Example LPR-MNIST target syntax samples} with random padding.}
\label{fig_lprmnist}
\end{figure}

\subsubsection{LPR-MNIST}
Genuine LP photographs come in a variety of image qualities and weather conditions which make models need larger datasets to learn to ignore irrelevant variations and cloud our main topic of interest, namely syntax modeling. Moreover, the real LP dataset is private as it contains non-disclosable personal data. We thereby create a proxy dataset offering several key advantages over real photographs, 
\begin{enumerate*}[label=(\roman*), itemjoin={{, }}, itemjoin*={{, and }}]
\item summarizing our key problem of syntax modeling 
\item discarding recognition issues arising from image degradation
\item accelerating experiments from using lower-resolution images and in lower numbers.
\end{enumerate*}

Let us introduce LPR-MNIST, a collection of 100,000 synthetic image-text pairs. It is easily generated by concatenating $L=5$ black-and-white digits from MNIST~\cite{lecunGradientbasedLearningApplied1998}, each padded to $32 \times 32$, giving $H = 32, W_\mathcal{I} = 160$. Here, $\Omega = \mathcal{D}$, the set of the 10 digits. Labels are drawn from \texttt{00000} to \texttt{99999} such that each possible combination is represented once. For each digit in a given label, the MNIST image is then randomly picked among all instances of this digit. Examples are given in Fig.~\ref{fig_lprmnist}. No particular measure was taken to reproduce the alphanumeric syntax, country indicators or rivets peculiar to real LP, as this simple design is sufficient to reproduce syntax shift recognition issues. Yet, as detailed in Section~\ref{expe}, sim-to-real transfer succeeds as our architecture modifications improve debiasing on both LPR-MNIST and real LP. This proxy dataset is meant to be used on its own, not for pretraining or as data augmentation for real LP. 

Creating a tipping point is as easy as purposely removing all samples starting with an arbitrary character at training time, here $\mathcal{N}_1 = \{ \texttt{9} \}$. Training and validation splits respectively contain 72,081 and 8,937 samples. Among the test split, 8,982 samples are of source syntax and 1,018 of target syntax.

\subsubsection{\label{frlpr}Real license plates}
The main goal of this paper is to reach production-ready text recognition accuracy on pre-localized cropped photographs of French LP captured in the wild. The corresponding dataset is composed of RGB photographs taken on the roads over a one-year period ranging from November 2020 to October 2021. For training purposes, annotations of the corresponding Latin alphanumeric letters and digits of length $L=7$ are readily available. Here, the character set is $\Omega = \mathcal{A} \cup \mathcal{D}$, where $\mathcal{A}$ is the set of alphabetic uppercase letters. 

As previously stated, the main difficulty which we address arises from the shift in the distribution of the characters on the LP as seen in Fig.~\ref{fig_frlpr_fail_trocr}. Due to the assignment of serial numbers in ascending order, the leftmost letter ultimately switches from one letter to the next, for example from an \texttt{F} to a \texttt{G}, thereby introducing a sudden mismatch between the trained model and the vehicle plates on the roads. Training and validation splits have their leftmost letter ranging from \texttt{A} to \texttt{F} and contain 797,469 and 97,725 samples, respectively. Some images starting with the letter \texttt{G} appeared around the end of 2021 are available in the test split, i.e. $\mathcal{N}_1 = \{ \texttt{G} \}$, so that 96,909 samples are of source syntax and 3,097 of target syntax. Each split contains images from the whole image capture period in order to dismiss any visual shift unrelated to syntax shift. 

\section{\label{expe}Experiments}

\subsection{Evaluation protocol}

\subsubsection{Exact match} 
To measure model performance, we compute the \textit{exact match} (also known as subset accuracy), a stricter version of accuracy in which all characters in a predicted string have to match their label counterpart exactly for the sample to be counted as correct. It is reported as a percentage.

\subsubsection{Seeds \& standard deviation} 
Despite being relevant for assessing the robustness of a method in bias reduction, standard deviation (SD) is rarely reported~\cite{huVisionLanguageAdaptiveMutual2023,resendegoncalvesRealTimeAutomaticLicense2018,trantienUnsupervisedDomainAdaptation2023,wanVocabularyRelianceScene2020a,yimSynthTIGERSyntheticText2021,zhangRelationalContrastiveLearning2023}. In this paper, all experiment are repeated on 5 random seeds so as to reliably report SD. Indeed, exact match must be analyzed through its mean and SD jointly since variability can be very high on target syntax, meaning that some models totally block the prediction of known characters in a new position. When analyzing results in our controlled experimental setting, we therefore aim for the lowest possible SD.

\subsection{Training details}
When not otherwise specified, models follow the training procedure described below. All models are trained or fine-tuned on the dataset on which test metrics are reported, namely LPR-MNIST or else real LP. Real LP images are resized to $H = 32, W_\mathcal{I} = 288$ before being given as input to the model so as to match the aspect ratio of LPR-MNIST as closely as possible. 32 pixels of zero-padding are randomly shared between the left and right sides of the image, hence $W = W_\mathcal{I} + 32$. While this padding was first designed for LPR-MNIST to mimic the natural variability in the absolute position of characters on real LP, we found that the latter benefit from it too. Text is processed by a character-level tokenizer. We use the standard cross-entropy loss. The optimizer is AdamW~\cite{loshchilovDecoupledWeightDecay2018} with weight decay equal to $10^{-4}$. Models are trained with a budget of 30 epochs and early stopping with a patience of 4 epochs. SaLT is always trained from scratch with a batch size of 512 and a learning rate of $10^{-4}$. TrOCR\textsubscript{SMALL} variant is used throughout this paper. For a fairer comparison, we replace TrOCR's byte-pair-encoding (BPE)~\cite{sennrichNeuralMachineTranslation2016} tokenizer with the character-level one used by SaLT. Indeed, as showcased in Table~\ref{table_frlpr_fail_trocr}, a character-level tokenizer largely outperforms BPE on some seeds, which we argue is due to BPE being prone to model inter-character relationships. When fine-tuning TrOCR, pretrained weights are loaded into the encoder while the decoder is randomly initialized to adapt to its newly defined character-level tokenizer. We start from the first-stage checkpoint\footnote{Available at \url{https://huggingface.co/microsoft}.} as it is meant for further fine-tuning. TrOCR uses a batch size of 256 and a learning rate of ${5 \times 10^{-5}}$. Its vision encoder keeps its own square image resizing strategy. 

\begin{table}
  \centering
  \begin{threeparttable}[c]
    \caption{\textbf{Performance comparison between vanilla and debiased model variants on real LP.} Exact match on target syntax leaps from mediocre to near-training-time performance with SaLT. Conversely, TrOCR remains strongly biased towards source syntax regardless of the decoder.}
    \label{table_debias_frlpr}
    \setlength{\tabcolsep}{4pt}
    \begin{tabular}{ccccc}
    \toprule
    \multirow{2}[1]{*}{\textbf{Model}}                                       & \multirow{2}[1]{*}{\textbf{Encoder}}             & \multirow{2}[1]{*}{\textbf{Decoder}}                      & \textbf{\makecell{Target \\ syntax}} & \textbf{\makecell{Source \\ syntax}} \\ & & & \verb/^G.*$/ & \verb/^[A-F].*$/ \\ \midrule
    TrOCR                           & \multirow{2}{*}{Transformer} & Transformer                           & $27.8\pm32.8$                          & $99.5\pm0.0$                           \\
    TrOCR\tnote{d} &                              & Transformer\tnote{d} & $63.3\pm34.1$                          & $99.5\pm0.0$                           \\ \hline \\[-2ex]
    FCN+Transformer                                      & \multirow{2}{*}{FCN}         & Transformer                           & $72.5\pm40.2$                          & $99.3\pm0.0$                           \\
    FCN+Transformer\tnote{d}~~(SaLT)                                                 &                              & Transformer\tnote{d} & $\bm{97.3\pm0.7}$                  & $99.3\pm0.0$                           \\ \bottomrule
    \end{tabular}
    \begin{tablenotes}
        \item [d] debiased.
    \end{tablenotes}
  \end{threeparttable}
\end{table}

\subsection{Results on real license plates}

\subsubsection{\label{salt_frlpr}SaLT is future-proof}

According to Table~\ref{table_debias_frlpr}, SaLT outperforms the standard TrOCR baseline by a significant margin on the real LP dataset, gaining +69.5 points on target syntax to reach 97.3\% mean exact match. Moreover, SD is reduced to a low 0.7\%, making any seed suitable for production. Meanwhile, peak performance is preserved on source syntax, nearing that of TrOCR. SaLT is thus future-proof albeit only using the biased dataset of past serials, retraining it at each change in LP syntax is therefore unnecessary. Additionally, we show that debiasing the decoder is key to future-proofing by comparing the two FCN-encoder variants. Indeed, with an FCN-encoder, switching from a biased Transformer-decoder to SaLT's decoder stabilizes target syntax performance while bringing a +24.8 points gain in mean exact match. 

\subsubsection{Transformer-encoders are disrupted by syntax shift}
For completeness, we also try our decoder-debiasing modifications on TrOCR. As reported in Table~\ref{table_debias_frlpr}, our attempt at debiasing TrOCR partially works with a significant improvement in mean exact match, although SD remains high for both the standard and debiased configurations. We argue it is impossible to debias TrOCR more efficiently because its Transformer-encoder can see the whole image and encode it with position information, hereby justifying the need for SaLT's position-less FCN-encoder with low receptive field $r$. 

\begin{table}
  \centering
  \begin{threeparttable}[c]
    \caption{\textbf{Ablation study of SaLT components on LPR-MNIST.} Here, a~{\xmark} denotes component removal among previous token reinjection, residual connection and position in $\bm{V}$, thus making a step towards debiasing.}
    \label{table_ablation_enc_dec_mnist}
    \setlength{\tabcolsep}{1pt}
    \begin{tabular}{ccccccc}
    \toprule
    \multirow{3}[6]{*}{\textbf{Model}}  & \multirow{3}[6]{*}{\textbf{Encoder}} & \multicolumn{3}{c}{\textbf{Decoder}}                                                                                                           & \multirow{2}[1]{*}{\textbf{\makecell{Target \\ syntax}}} & \multirow{2}[1]{*}{\textbf{\makecell{Source \\ syntax}}} \\ \cline{3-5}
                                     &                                   & \multirow{2}{*}{\textbf{\makecell{Reinject \\ previous \\ token}}} & \multicolumn{2}{c}{Last layer}                                           &                                                       &                                                       \\ \cline{4-5}
                                     &                                   &                                                                     & \textbf{Residual} & \makecell{\textbf{Position} \\ \textbf{in} $\bm{V}$} & \verb/^9.*$/                                                      & \verb/^[0-8].*$/                                                      \\ \midrule
    TrOCR       & Transformer                       & \vmark                                                              & \vmark            & \vmark                                               & $85.6\pm7.3$                                          & $98.8\pm0.4$                                          \\ \hline \\[-2ex]
    \multirow{7}{*}{FCN+Transformer} & \multirow{7}{*}{FCN}              & \vmark                                                              & \vmark            & \vmark                                               & $87.1\pm8.4$                                          & $99.3\pm0.1$                                          \\
                                     &                                   & \vmark                                                              & \vmark            & \xmark                                               & $90.8\pm6.7$                                          & $99.3\pm0.1$                                          \\
                                     &                                   & \vmark                                                              & \xmark            & \vmark                                               & $93.9\pm2.6$                                          & $99.4\pm0.1$                                          \\
                                     &                                   & \vmark                                                              & \xmark            & \xmark                                               & $94.5\pm1.9$                                          & $99.3\pm0.1$                                          \\
                                     &                                   & \xmark                                                              & \vmark            & \vmark                                               & $95.5\pm0.7$                                          & $99.3\pm0.1$                                          \\
                                     &                                   & \xmark                                                              & \vmark            & \xmark                                               & $95.1\pm1.0$                                          & $99.3\pm0.1$                                          \\
                                     &                                   & \xmark                                                              & \xmark            & \xmark                                               & $95.7\pm1.5$                                          & $99.4\pm0.1$                                          \\ \hline \\[-2ex]
    \makecell{FCN+Transformer\tnote{d} \\ (SaLT)}                             & FCN                               & \xmark                                                              & \xmark            & \vmark                                               & $\bm{96.6\pm0.9}$                                     & $99.4\pm0.1$                                          \\ \bottomrule
    \end{tabular}
    \begin{tablenotes}
        \item [d] debiased.
    \end{tablenotes}
  \end{threeparttable}
\end{table}

\subsection{\label{ablation}Ablation study on LPR-MNIST}

\subsubsection{\label{mnist_ablation}Joint effect of SaLT components}
In Table~\ref{table_ablation_enc_dec_mnist}, we present results after ablating critical components of SaLT on LPR-MNIST, including TrOCR reaching a baseline of 85.6\% mean exact match with 7.3\% SD. We observe an overall trend of increase in mean exact match and decrease in SD, proving that our debiasing proposals tend to positively complement each other when combined. Meanwhile, exact match on source syntax is not altered. 

Let us now consider each decoder modification separately, starting from an FCN-encoder with a biased decoder. Preferring position queries to previous token reinjection gives the highest gains on target syntax, with a biggest leap of +8.4 points mean and -7.7 points SD. Also, it consistently improves both the mean and SD compared to previous-token-reinjection counterparts. Cutting off the cross-attention residual is the second most consistent modification, with a biggest leap of +6.8 points mean and -5.8 points SD. When combined with position queries however, the gain in mean comes at the cost of a minor increase in instability between seeds. Lastly, while leaving $\bm{V}$ without positional encoding does bring a highest gain of +3.7 points mean and -1.7 points SD, combining it with position queries leads to small performance drops. 

Consequently, the overall best configuration chosen for SaLT comprises only two of the three architectural modifications. We notice that removing previous token reinjection is the most direct way of cancelling language modeling. Cutting off the cross-attention residual further brings a small performance gain. 

\begin{table}
  \centering
  \begin{threeparttable}[c]
    \caption{\textbf{Performance comparison between vanilla and debiased model variants on LPR-MNIST without encoder dropout.} Decoder debiasing on LPR-MNIST also works when encoder dropout is disabled.}
    \label{table_ablation_dropout}
    \setlength{\tabcolsep}{28pt}
    \begin{tabular}{ccc}
    \toprule
    \multirow{2}[1]{*}{\textbf{Model}}                        & \textbf{\makecell{Target \\ syntax}} & \textbf{\makecell{Source \\ syntax}} \\ & \verb/^9.*$/ & \verb/^[0-8].*$/ \\ \midrule
    FCN+Transformer                        & $89.1\pm7.1$             & $97.6\pm0.6$             \\
    FCN+Transformer\tnote{d}     & $\bm{95.0\pm1.4}$             & $98.4\pm0.2$             \\ \bottomrule
    \end{tabular}
    \begin{tablenotes}
        \item [d] debiased.
    \end{tablenotes}
  \end{threeparttable}
\end{table}

\subsubsection{Impact of dropout}

Assuming that dropout should help alleviating bias for being a regularization method, we further report debiasing performance without dropout in the FCN on LPR-MNIST. Table~\ref{table_ablation_dropout} demonstrates that decoder debiasing without encoder dropout brings a gain on target syntax close to that with it enabled (previously shown in Table~\ref{table_ablation_enc_dec_mnist}). Yet, enabling dropout achieves better results both on source and target syntax.

\begin{table}
  \centering
  \begin{threeparttable}[c]
    \caption{\textbf{Performance comparison between vanilla and debiased model variants on LPR-MNIST with digit \texttt{9} missing in various positions.} All missing character positions benefit from SaLT's decoder debiasing.}
    \label{table_ablation_posx}
    \setlength{\tabcolsep}{12pt}
    \begin{tabular}{cccc}
    \toprule
    \multirow{2}[1]{*}{\textbf{Model}}                                                         & \multirow{2}[1]{*}{\textbf{Position}} & \textbf{\makecell{Target \\ syntax}} & \textbf{\makecell{Source \\ syntax}} \\ & & \verb/^9.*$/ & \verb/^[0-8].*$/ \\ \midrule
    \multirow{4}{*}{FCN+Transformer}                         & 2                 & $95.3\pm1.3$             & $99.2\pm0.1$             \\
                                                                                             & 3                 & $95.4\pm5.1$             & $99.2\pm0.1$             \\
                                                                                             & 4                 & $80.2\pm35.2$            & $99.2\pm0.1$             \\
                                                                                            & 5                 & $93.2\pm2.0$             & $99.2\pm0.1$             \\ \hline \\[-2ex]
    \multirow{4}{*}{FCN+Transformer\tnote{d}~~(SaLT)}             & 2                 & $\bm{96.4\pm0.8}$             & $99.3\pm0.1$             \\
                                                                                             & 3                 & $\bm{97.6\pm0.9}$             & $99.3\pm0.1$             \\
                                                                                             & 4                 & $\bm{98.3\pm0.7}$             & $99.3\pm0.1$             \\
                                                                                             & 5                 & $\bm{96.7\pm1.0}$             & $99.3\pm0.1$             \\ \bottomrule
    \end{tabular}
    \begin{tablenotes}
        \item [d] debiased.
    \end{tablenotes}
  \end{threeparttable} 
\end{table}

\subsubsection{Generalization to other positions}

We investigate digit \texttt{9} missing in other positions in LPR-MNIST, i.e. $\mathcal{N}_i = \{ \texttt{9} \}, \, i \in \{2,3,4,5\}$. SaLT achieves debiasing on all positions. While position $i = 4$ gains +18.1 points mean and -34.5 points SD, others have decent initial exact match leaving little margin for improvement. 

\section{\label{conclu}Conclusion \& future work}
This work investigates the robustness to syntax shift of Transformers for license plate recognition. After showing the role of contextual and positional information flows in syntax modeling, we introduce SaLT, a new syntax-less Transformer architecture. SaLT tackles training-time syntax absorption by using an FCN-encoder and cutting off key components in an enhanced Transformer-decoder. Distinct from popular attention-based networks hallucinating on legible images, it exhibits future-proof recognition performance on real-life vehicle plates and synthetic datasets. Indeed, extensive experiments show that SaLT achieves high recognition accuracy and low variability on future syntax while fully preserving performance on past syntax. Moreover, it provides competitive debiasing in any position on a license plate. We are therefore optimistic about its robustness to other characters or other countries with their own syntax. Better understanding the receptive field of FCN-encoded features and the impact of supplying them with positional information may be key to bridging the small performance gap separating target and source syntax. We also note the very goal of relaxing syntax modeling comes at the cost of not learning parts of the syntax which could be useful for prediction, e.g. the ordering between letters and digits in licenses, hence future work will concentrate on re-injecting desirable constraints.

\begin{credits}
\subsubsection{\ackname} This study was funded by the CIFRE ANRT grant No. 2023/0195.
\end{credits}

\bibliographystyle{splncs04}
\bibliography{biblio}

\end{document}